\documentclass[10pt,twocolumn,letterpaper]{article}

\usepackage{iccv}
\usepackage{times}
\usepackage{epsfig}
\usepackage{graphicx}
\usepackage{amsmath}
\usepackage{amssymb}
\usepackage{mathtools}
\usepackage{amsthm}
\usepackage{bbm}
\usepackage{paralist}
\usepackage[title]{appendix}

\usepackage[textsize=tiny]{todonotes}

\usepackage[pagebackref=true,breaklinks=true,letterpaper=true,colorlinks,bookmarks=false]{hyperref}

\iccvfinalcopy 

\ificcvfinal\fi

\begin{document}

\title{Practical Membership Inference Attacks Against Large-Scale Multi-Modal Models: A Pilot Study}

\author{Myeongseob Ko$^1$\hspace{1.5em}
Ming Jin$^1$\hspace{1.5em}
Chenguang Wang$^2$\hspace{1.5em}
Ruoxi Jia$^1$\\[1ex] 
$^1$Virginia Tech\hspace{1.5em} $^2$Washington University in St. Louis\\
{\tt\small \{myeongseob, jinming, ruoxijia\}@vt.edu, chenguangwang@wustl.edu}
}


\maketitle
\ificcvfinal\thispagestyle{empty}\fi

\begin{abstract}
    Membership inference attacks (MIAs) aim to infer whether a data point has been used to train a machine learning model. These attacks can be employed to identify potential privacy vulnerabilities and detect unauthorized use of personal data. While MIAs have been traditionally studied for simple classification models, recent advancements in multi-modal pre-training, such as CLIP, have demonstrated remarkable zero-shot performance across a range of computer vision tasks. However, the sheer scale of data and models presents significant computational challenges for performing the attacks. 
    
    This paper takes a first step towards developing practical MIAs against large-scale multi-modal models. We introduce a simple baseline strategy by thresholding the cosine similarity between text and image features of a target point and propose further enhancing the baseline by aggregating cosine similarity across transformations of the target. We also present a new weakly supervised attack method that leverages ground-truth non-members (e.g., obtained by using the publication date of a target model and the timestamps of the open data) to further enhance the attack. Our evaluation shows that CLIP models are susceptible to our attack strategies, with our simple baseline achieving over $75\%$ membership identification accuracy. Furthermore, our enhanced attacks outperform the baseline across multiple models and datasets, with the weakly supervised attack demonstrating an average-case performance improvement of $17\%$ and being at least $7$X more effective at low false-positive rates. These findings highlight the importance of protecting the privacy of multi-modal foundational models, which were previously assumed to be less susceptible to MIAs due to less overfitting. Our code is available at https://github.com/ruoxi-jia-group/CLIP-MIA.
\end{abstract}

\section{Introduction}
\label{introduction}

Membership inference attack (MIA) is a type of privacy attack that attempts to determine if a specific data point was used to train a machine learning model \cite{hu2022membership}. This type of attack can compromise the privacy of individuals whose data was used to train the models \cite{hu2022membership}, but can be also used to identify vulnerabilities, privacy leakage, and unauthorized use of personal data in machine learning models \cite{nasr2021adversary,song2019auditing}.

Moreover, the rise of a foundational model trained on vast amounts of open data has highlighted the potential breach of contextual integrity, a fundamental principle in legal discussions of privacy \cite{nissenbaum2009privacy}. MIA can, therefore, be used as an effective tool for individuals to check if companies store their personal information and request its deletion to comply with the European General Data Protection Regulation (GDPR).


Existing MIAs that achieve advanced performance rely on the idea of shadow training ~\cite{carlini2022membership,wen2022canary}. The shadow models are usually at a scale of hundreds~\cite{carlini2022membership} and are with the same or similar architecture as the target model.
The training algorithm needs to be the \emph{same} as the one that trains the target model. 
The difference between shadow models that contain a certain sample and those that do not is then utilized to learn proper features to identify the membership of an individual sample. However, reliance on shadow training forecloses its application to large-scale models. As a concrete example, training Contrastive Language-Image Pre-Training (CLIP)~\cite{radford2021learning}, a cutting-edge multi-modal learning paradigm, takes 18 days even with hundreds of advanced GPUs~\cite{guwukong}. This makes shadow training to attack CLIP prohibitively expensive. In addition, obtaining full details of the training algorithm performed by these state-of-the-art models is often difficult, as they are considered intellectual property and are not published.

This paper is a pilot study of practical MIAs against large-scale multi-modal models. The proposed techniques bypass the shadow training and work with only black-box access. We use CLIP as an example for the extensive evaluation, due to the following reasons: First, CLIP is widely used for zero-shot learning in various computer vision tasks~\cite{ramesh2021zero,ramesh2022hierarchical,conde2021clip}, which entails the understanding of its privacy risk. 
Also, CLIP has been trained on massive data scraped from the Internet with undisclosed algorithms, thus exemplifying the scale and the threat model for other emergent foundation models~\cite{bommasani2021opportunities}. Our \underline{contributions} are summarized as follows.


\textbf{(1) Benchmarking the susceptibility of CLIP to membership inference.} We introduce a simple baseline attack that identifies membership based on cosine similarity (CS) between image and text features, akin to those MIAs designed for single-modal models based on loss or confidence scores in prior literature~\cite{yeom2018privacy,leino2020stolen,song2020systematic}. The rationale for using CS as a signal for membership inference is that CLIP is trained to maximize CS on training samples, which can result in members having higher CS than non-members. This simple method achieves a reasonable membership identification accuracy ranging from $66.5\%$ to $78.8\%$ but has limited performance in the low false-positive regime. 

\textbf{(2) Improving the CS-based attack via target data augmentations.} We develop an enhanced MIA technique that involves applying transformations to a specific point and aggregating the changes in CS across various common transformations (e.g., resizing, cropping, rotation, and translation). Our evaluation indicates that this method consistently improves attack performance across multiple datasets and CLIP model architectures, albeit to a small extent. The inspiration for this technique stems from our empirical observations that training points experience a larger CS drop than non-training ones when subject to transformations. 

\textbf{(3) Developing a new weakly supervised MIA framework given one-sided non-member information.} Both the baseline attack and the augmentation-enhanced attack demonstrate high performance without requiring ancillary information, but their accuracy is limited at false positive rates (e.g., less than $10\%$ true positive rate at a false positive rate of $1\%$). In this paper, we identify a new threat model that is plausible for models trained on Internet data. Specifically, we consider a scenario where the attacker has \emph{one-sided} knowledge about non-members. By scraping Internet data posted after the target model's publication date, the attacker can acquire a set of data guaranteed not to have participated in the training. We propose a weakly supervised attack that utilizes this non-member set to construct a model that predicts membership. This approach shows remarkable attack performance, improving the baseline accuracy by $17\%$  and being around $7$X more effective at low false-positive rates, despite the absence of information about members.

\textbf{(4) Exploring potential defenses.} Our findings demonstrate the vulnerability of large-scale multi-modal models to membership inference risks. To address this issue, we investigate potential defenses and their associated privacy-utility tradeoff.
This pilot study serves as a starting point for assessing the privacy risks associated with emerging large-scale multi-modal models.

\section{Related Work}
\label{related work}

Membership inference attacks (MIAs) are designed to determine whether a given data sample has been used to train a particular model. These attacks typically leverage the overfitting tendency of machine learning models, which often show varying predictive behaviors on training data versus unseen data. While most existing MIAs have been developed for small-scale classification models~\cite{hu2022membership}, where overfitting is more prevalent, this paper focuses on investigating MIAs against large-scale multi-modal models.

One approach taken by MIAs involves identifying a metric that can differentiate between the behaviors of member and non-member samples. Example metrics include prediction loss~\cite{yeom2018privacy}, correctness~\cite{leino2020stolen,song2020systematic}, and entropy~\cite{shokri2017membership,song2020systematic}. 
Our baseline attack strategy is similar to the metric-based attack, adapted to the multi-modal setting.

Another popular approach~\cite{shokri2017membership} is through building an attack model, which takes the predicted confidence vector of the target model on a target sample as input and infers the membership of the sample as output. To obtain the ``training data'' for the attack model, one needs to perform shadow training, which involves training numerous models on different subsets of data. A point's membership within a shadow model and the corresponding features extracted from that shadow model can then form a training pair for the attack model.
Recent studies by \cite{carlini2022membership,wen2022canary} have explored likelihood-ratio-based MIAs, which estimate the probability of having a certain loss for both members and non-members and then identify the membership based on the likelihood ratio. These approaches also require shadow training.
The losses of the shadow models trained with and without the target point are used to estimate the likelihood ratio. Although these techniques achieve state-of-the-art attack performance, especially in the low false-positive regime, their reliance on shadow training makes them computationally infeasible for large-scale models. Additionally, shadow training assumes knowledge of the training procedure and target model architecture, which becomes problematic as many large models are only available as black-box APIs.


MIAs against text-to-image generation models~\cite{wu2022membership} and image captioning models~\cite{hu2022m} have recently been proposed to keep pace with the increasing use of multi-modal models. While sharing a similar scope to our work, their considered training size is relatively small, and thus shadow training is still applicable. By contrast, we propose MIAs for concurrent large-scale multi-modal models for which shadow training is intractable. A recent paper~\cite{hintersdorf2022clipping} also studies privacy risks associated with CLIP; however, unlike our paper which provides a general, application-agnostic attack method, their method can only be applied to inferring whether an identity's face images are in the training set. 

\section{Threat Model}
\label{background}

This paper focuses on designing MIAs against large-scale vision-language models trained via CLIP~\cite{radford2021learning}.
CLIP consists of an image encoder $f_\text{img}$ and a text encoder $f_\text{txt}$. Given a training set $\mathcal{D}_\text{trn}=\{(x_i,y_i)\}_{i=1,...,n}$, where $x_i$ represents the image portion of the $i$-th training sample and $y_i$ represents the corresponding text portion. CLIP jointly trains the image encoder and the text encoder 
via contrastive learning. The training objective aims to maximize the CS between correct pairs of image and text features while minimizing the similarity between incorrect pairs.

We summarize the attacker's goal and the assumptions on the attacker's knowledge as follows.

\vspace{-0.5em}
\paragraph{Goal.} Consider the training phase of a target model, samples in the training set are called \emph{members}, and those not used for training are called \emph{non-members}. For a target model, the goal of a membership inference attack (MIA) is to infer whether a given sample is a member or not.

\vspace{-0.5em}
\paragraph{Model Knowledge.} We consider \emph{black-box} access to the target model, where the attacker can query the model with pairs of images and corresponding text captions, and in return receive image and text features. In particular, the attacker does not know the architectural design and parameters of the target model or the training algorithm.

\vspace{-0.5em}
\paragraph{Data Knowledge.} We explore two different settings regarding the attacker's knowledge of the training data used for the target model. In the first setting, called the \emph{zero-knowledge setting}, we assume that the attacker has no knowledge of the data distribution used to train the target model. In the second setting, which is similar to past research~\cite{carlini2022membership,wen2022canary,shokri2021privacy}, we assume that the attacker has access to a set of samples $\mathcal{D}_\text{all}$ that represents the underlying data distribution $\mathbb{D}$. This assumption is based on the fact that large-scale multi-modal models are often trained on data collected from the internet, allowing the attacker to scrape data from the internet as a proxy for the underlying distribution. In this second setting, we further assume that a set $\mathcal{D}_\text{no}$ is known to be non-members, while the membership of $\mathcal{D}_\text{all}$ is unknown. For example, one can easily collect non-members by scraping the data posted after the model's publication, but membership of data posted before the publication date is not discernible. This indicates that ground-truth membership information is \emph{one-sided}: we can only obtain information about non-members precisely. More specifically, consider the open-CLIP models, which are trained on LAION 400M. The authors have released information regarding their data collection: ‘The dataset has been extracted from random web pages crawled between 2014 and 2021.’ Furthermore, the original CLIP model was published in January 2021, and GPT-4 was officially launched on March 2023, training on data collected before September 2021. Therefore, it is realistic for prospective attackers to leverage the published date of the target model and create a non-member dataset by scraping the Internet data posted after the publication date of the target model.
Alternatively, one can obtain a ground-truth set of non-members by keeping some data internal and not contributing it to the training of the target model. We refer to this second setting as the \emph{one-sided-knowledge setting}.

\vspace{-0.5em}
\paragraph{Computational Resources.} 
Our focus is on a practical attack scenario, where the attacker has limited computing resources and must rely on efficient attack strategies. Specifically, our goal is to entirely circumvent shadow training computations, which involve repeatedly re-training the target model.

\section{Methodology}
In this section, we introduce three attack strategies for large-scale multi-modal models. The first two strategies are applicable to the zero-knowledge setting, while the third strategy can take advantage of one-sided non-member information.

\subsection{A Baseline: Cosine Similarity Attack}
The key idea is that the target model is trained to maximize cosine similarity between image and text features on members, so the attacker will receive higher cosine similarity scores from members than from non-members. Given the cosine similarity between the input image $x$ and the text $y$, we predict $(x,y)$ as a member if $\texttt{CS}(f_\text{img}(x),f_\text{txt}(y))>\tau$ for some threshold $\tau>0$.
We refer to this attack as the \emph{Cosine Similarity Attack (CSA)}. This attack has a similar flavor to those that threshold classification confidence score or loss to infer data membership for classification models~\cite{yeom2018privacy}. 

\subsection{Augmentation-Enhanced Attack}
\label{sec:AEA}

One idea for achieving stronger attack performance is to leverage the data augmentation technique, inspired by ~\cite{kaya2021does,choquette2021label}. In particular, for multi-modal models, we observe that after applying transformations to each target sample, the decrease in cosine similarity is more significant for member samples than for non-member samples. This observation enables us to leverage the CS gap induced by transformations to improve the attack performance. 

Formally, we consider $K$ transformations $\mathcal{T}=\{T_1(\cdot),\ldots,T_K(\cdot)\}$. For each transformation, we compute the CS associated with the transformed target sample, namely $\texttt{CS}(f_\text{img}(T_k(x)),f_\text{txt}(y))$, and then calculate the gap from the CS associated with the original sample:
\begin{align}
\label{eqn:transform_gap}
&\Delta \texttt{CS}_k (x,y) \nonumber\\
&:=\texttt{CS}(f_\text{img}(x),f_\text{txt}(y)) - \texttt{CS}(f_\text{img}(T_k(x)),f_\text{txt}(y)).
\end{align}
To improve the baseline attack, we aggregate CS gaps over all transformations considered in addition to the CS of the original image, i.e., $\texttt{CS}(x,y) + \sum_{k=1}^K \Delta\texttt{CS}_k(x,y)$, and predict $(x,y)$ as a member when this aggregate quantity exceeds a certain threshold. We refer to this method as the \emph{Augmentation-Enhanced Attack (AEA)}.



\subsection{Weakly Supervised Attack}

\begin{figure}[h!]
\centering
\includegraphics[width=1.0\linewidth]{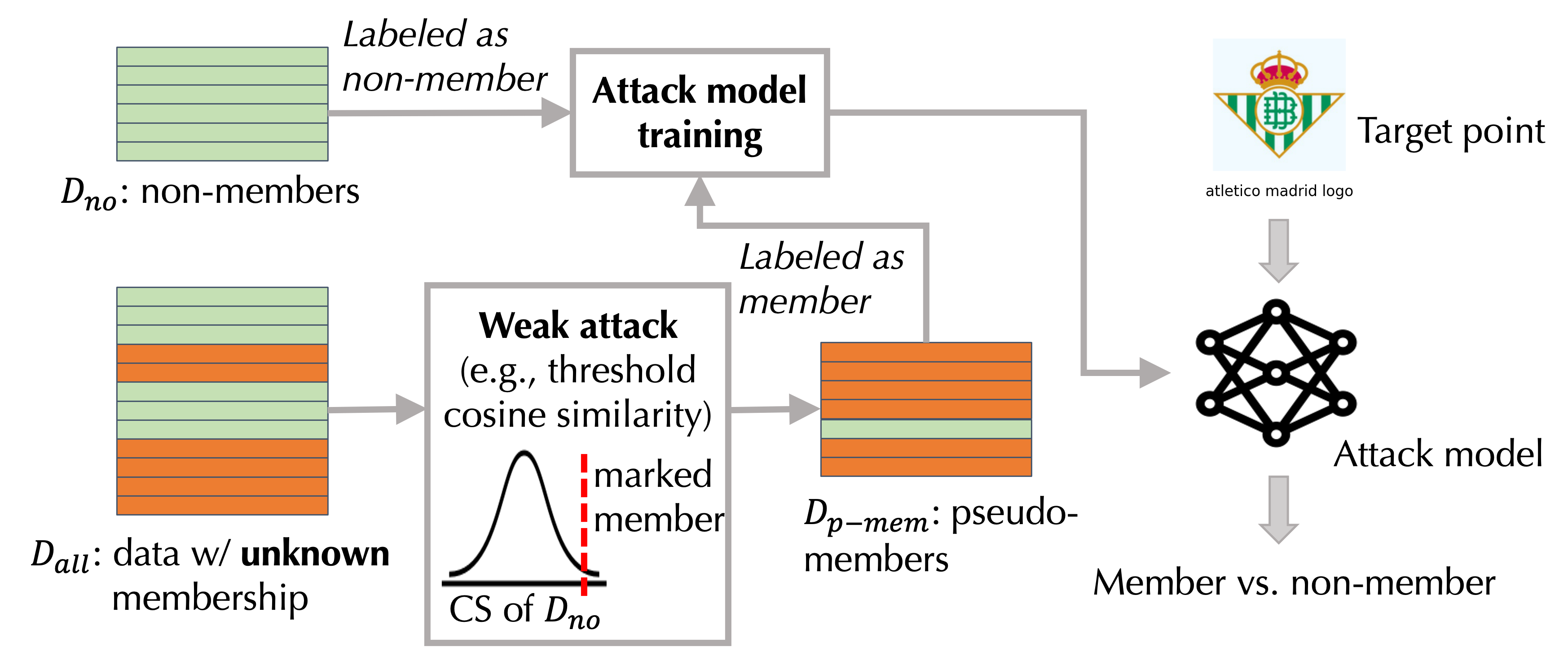}
\caption{Overview of the WSA.}
\label{fig:overview}
\end{figure}

Different from the AEA, our third approach uses ground-truth information about non-members \emph{in tandem with} noisy information about members provided by a \emph{weak} attack method to learn an attack model. The attack pipeline is illustrated in Figure~\ref{fig:overview}.

The attack starts by querying the target model for the image and text features associated with each non-member in $\mathcal{D}_\text{no}$. We denote the collection of non-member features as $F_\text{no}=\{(f_\text{img}(x),f_\text{txt}(y))|(x,y)\in \mathcal{D}_\text{no}\}$. At the same time, we evaluate CS between the features: $CS_\text{no}=\{\texttt{CS}(f_\text{img}(x),f_\text{txt}(y))|(x,y)\in \mathcal{D}_\text{no}\}$. Based on ${CS}_\text{no}$, we can estimate the distribution of CS scores for non-members. As we will show later, the distribution has a Gaussian shape. We can further approximate the mean and variance of the distribution through the sample mean and sample variance of $CS_\text{no}$, denoted by $\mu_\text{no}$ and $\sigma^2_\text{no}$. Intuitively, a sample with CS significantly larger than $\mu_\text{no}$ (e.g., larger by $\lambda\sigma_\text{no}$ for some constant $\lambda$) is likely to be a member.

Now, our goal is to find a subset of $\mathcal{D}_\text{all}$ most likely to be members. To achieve this, we query the target model for image-text features associated with $\mathcal{D}_\text{all}$ and calculate the CS score for each sample. We compare the scores of each sample with the distribution of non-member scores characterized by $\mu_\text{no}$ and $\sigma_\text{no}$ and further mark the samples with CS significantly larger than $\mu_\text{no}$ as members. We call the set of members marked based on CS the \emph{pseudo-member set} because this set is likely to contain non-members as well. Formally, we represent the pseudo-member set as 
{\small
\begin{align}
    \mathcal{D}_\text{p-mem}=\{(x,y)|\texttt{CS}(x,y)\geq\mu_\text{no}+\lambda \sigma_\text{no},(x,y)\in \mathcal{D}_\text{all}\}.
\end{align}
}

Note that the pseudo-member set $\mathcal{D}_\text{p-mem}$ only provides noisy information about membership.

Given $\mathcal{D}_\text{no}$ and $\mathcal{D}_\text{p-mem}$, we can construct an \emph{attack dataset} by associating each text-image feature derived from $\mathcal{D}_\text{no}\cup \mathcal{D}_\text{p-mem}$ with a label indicating membership. Specifically, the attack dataset is denoted by
\begin{align}
    \mathcal{D}_\text{attack}=\{(f_\text{img}(x),f_\text{txt}&(y),b)|(x,y)\in \mathcal{D}_\text{no}\cup \mathcal{D}_\text{p-mem}, \nonumber\\
    &b=\mathbbm{1}[(x,y)\in \mathcal{D}_\text{p-mem}]\},
\end{align}
where $b$ is binary with $1$ representing a potential member and $0$ representing a potential non-member.

Finally, using $\mathcal{D}_\text{attack}$, we can train an attack model $f_\text{attack}$ that takes the image-text as input and returns its membership status. Once the attack model is trained, we can use it to attack any target sample $(x,y)$ by first querying the target model for its features $(f_\text{img}(x),f_\text{txt}(y))$ and feeding the features at the input of $f_\text{attack}$ to receive the membership inference result, namely, $f_\text{attack}(f_\text{img}(x),f_\text{txt}(y))$.

\section{Evaluation}
\label{evaluation}
In this section, we thoroughly evaluate the performance of our attacks (i.e., CSA, AEA, and WSA) across various models and datasets and explore potential defenses.


\begin{table*}[!h]
\caption{Attack performance evaluation across different target model architectures and datasets. $\Delta$ indicates the improvement from CSA.}
\label{tab:main_laion}
\centering
\renewcommand{\arraystretch}{1.7}
\scalebox{0.61}{
\begin{tabular}{cccccccccccccc}
\hline
Dataset &
  \multicolumn{13}{c}{LAION} \\ \hline
Model &
  ViT-B/32 &
   &
   &
   &
  ViT-B/16 &
   &
   &
   &
  ViT-L/14 &
   &
   &
   &
  \multicolumn{1}{l}{Runtime} \\ \hline
Metric &
  AUC &
  $\Delta$ &
  TPR@1\%FPR &
  $\Delta$ &
  AUC &
  $\Delta$ &
  TPR@1\%FPR &
  $\Delta$ &
  AUC &
  $\Delta$ &
  TPR@1\%FPR &
  $\Delta$ &
  (Sec) \\ \hline
CSA &
  0.7460 $\pm$ 0.0012 &
  - &
  0.0723 $\pm$ 0.0050 &
  - &
  0.7593 $\pm$ 0.0117 &
  - &
  0.0758 $\pm$ 0.0004 &
  - &
  0.7876 $\pm$ 0.0031 &
  - &
  0.0487 $\pm$ 0.0037 &
  - &
  670.7s \\
AEA &
  0.7625 $\pm$ 0.0012 &
  \textbf{0.0165} &
  0.0940 $\pm$ 0.0004 &
  \textbf{0.0217} &
  0.7897 $\pm$ 0.0003 &
  \textbf{0.0304} &
  0.0920 $\pm$ 0.0005 &
  \textbf{0.0162} &
  0.7950 $\pm$ 0.0025 &
  \textbf{0.0074} &
  0.0836 $\pm$ 0.0023 &
  \textbf{0.0349} &
  830.1s \\
WSA &
  0.9199 $\pm$ 0.0009 &
  \textbf{0.1739} &
  0.7212 $\pm$ 0.0069 &
  \textbf{0.6489} &
  0.9349 $\pm$ 0.0044 &
  \textbf{0.1756} &
  0.7381 $\pm$ 0.0061 &
  \textbf{0.6623} &
  0.9413 $\pm$ 0.0026 &
  \textbf{0.1537} &
  0.7611 $\pm$ 0.0104 &
  \textbf{0.7124} &
  1539.2s \\ \hline
Dataset &
  \multicolumn{13}{c}{CC12M} \\ \hline
Model &
  RN50 &
   &
   &
   &
  RN101 &
   &
   &
   &
  ViT-B/32 &
   &
   &
   &
  Runtime \\ \hline
Metric &
  AUC &
  $\Delta$ &
  TPR@1\%FPR &
  $\Delta$ &
  AUC &
  $\Delta$ &
  TPR@1\%FPR &
  $\Delta$ &
  AUC &
  $\Delta$ &
  TPR@1\%FPR &
  $\Delta$ &
  (Sec) \\ \hline
CSA &
  0.6650 $\pm$ 0.0026 &
  - &
  0.0224 $\pm$ 0.0011 &
  - &
  0.6854 $\pm$ 0.0040 &
  - &
  0.0361 $\pm$ 0.0011 &
  - &
  0.6957 $\pm$ 0.0051 &
  - &
  0.0395 $\pm$ 0.0013 &
  - &
  670.7s \\
AEA &
  0.7447 $\pm$ 0.0029 &
  \textbf{0.0797} &
  0.0948 $\pm$ 0.0039 &
  \textbf{0.0724} &
  0.7306 $\pm$ 0.0000 &
  \textbf{0.0453} &
  0.0930 $\pm$ 0.0020 &
  \textbf{0.0569} &
  0.705 $\pm$ 0.0016 &
  \textbf{0.0093} &
  0.0775 $\pm$ 0.0013 &
  \textbf{0.0380} &
  830.1s \\
WSA &
  0.7945 $\pm$ 0.0066 &
  \textbf{0.1295} &
  0.3724 $\pm$ 0.0138 &
  \textbf{0.3500} &
  0.8171 $\pm$ 0.0031 &
  \textbf{0.1318} &
  0.4032 $\pm$ 0.0066 &
  \textbf{0.3671} &
  0.7931 $\pm$ 0.0006 &
  \textbf{0.0974} &
  0.3306 $\pm$ 0.0058 &
  \textbf{0.2911} &
  1539.2s \\ \hline
\end{tabular}
}
\end{table*}

\subsection{Evaluation Setup}
\label{sec:main_data}
\paragraph{Datasets.} To evaluate our attack, we use the mixture of LAION~\cite{schuhmann2021laion}, Conceptual Captions 3M (CC3M) ~\cite{sharma2018conceptual}, Conceptual Captions 12M (CC12M) ~\cite{changpinyo2021conceptual}, and MSCOCO~\cite{lin2014microsoft} datasets as $\mathcal{D}_\text{all}$. These datasets are all in web-data format (e.g., image URLs and corresponding text captions) and can be scraped using the technique in~\cite{beaumont-2021-img2dataset}. LAION, CC3M, CC12M, and MSCOCO contain approximately 400M, 3.3M, 12M, and 600K image-text pairs, respectively. In particular, the LAION dataset is used by the open-sourced CLIP (OpenCLIP) implementation that we consider in this paper. We use one of the datasets to train a target model and sample from the combination of the other three datasets to get the ground-truth non-members. We further split the non-members into two disjoint subsets: one provides the ground-truth non-member information for the WSA, while the other is used for evaluating the attack performance. 

Note that our setup of using different open-world datasets as members and non-members differs slightly from the traditional MIA evaluation setting focused on small-scale classification models, where the same dataset is split into two subsets as members and non-members. In the main paper, we did not adopt the traditional setting for two reasons. Firstly, all the datasets considered are open-world data collected from the Internet; our setting, where the model owner uses one dataset to train the model and the attacker scrapes another dataset as non-member, simulates the real-world attack case. Secondly, the original OpenCLIP is trained on the \emph{full} LAION dataset and we would like to attack this model to demonstrate the real-world impact of our attacks. In fact, we found that training OpenCLIP on smaller subsets of LAION (e.g., 240M) leads to significant performance degradation, thereby yielding a less interesting attack case. 


\vspace{-0.5em}
\paragraph{Data Processing.}
For large-scale datasets scraped from the internet at scale, it is natural to have some overlapping (e.g., LAION has overlapped samples with other datasets). Therefore, we include the text preprocessing step and the URL processing step to remove the intersection so that $ \mathcal{D}_\text{no} \cap D_\text{trn} = \emptyset $ and the image-text pairs used to build $ D_\text{attack}$ has no overlap with those used to evaluate the attack performance. The details of the preprocessing step are provided in Appendix ~\ref{sec:F}. 

\vspace{-0.5em}
\paragraph{Target Models.}
Since the original CLIP paper~\cite{radford2021learning} does not disclose their training algorithm details, we utilize the reproduced OpenCLIP~\cite{cherti2022reproducible} as a surrogate, and it achieves performance comparable to what was reported in the original paper.
For the LAION dataset, we primarily focus on a vision transformer-based architecture (ViT)~\cite{dosovitskiy2020image}, given its superior performance. We utilize ViT-B/32, ViT-B/16, and ViT-L/14 as our target pre-trained models, following~\cite{cherti2022reproducible}.
We use a pre-trained ResNet50 (RN50) trained on the CC12M dataset and additionally train ResNet101 (RN101) and ViT-B/32 on the same dataset to evaluate our attack performance on different datasets and architectures. We will refer to ViT-B/32 and RN101 on CC12M as self-trained models, while the others will be referred to as pre-trained models. 
It is worth noting that ResNet structures are recommended for CC12M since vision transformers tend to be more data-hungry than ResNet~\cite{he2015deep}. 

\vspace{-0.5em}
\paragraph{Evaluation Metrics.}
We evaluate our approach based on three metrics. The first is the area under the curve (AUC) score of the receiver operating characteristic (ROC) curve. This is an average metric that takes an average over all false-positive rates. The second one is TPR@1\%FPR (true positive rate when the false positive rate is 1\%) which was used by ~\cite{carlini2022membership, wen2022canary} as a more practical measure since high FPR is not desirable for the attacker. Moreover, for WSA, we also consider the accuracy (ACC) of the attack model as a third metric, presented in Appendix ~\ref{sec:B}.


\subsection{Evaluation of Proposed Attacks}
\label{sec:main_eval}

\paragraph{Attack Performance Evaluation.} We evaluate the attack performance of our proposed attacks, namely, CSA, AEA, and WSA, on target models. AEA utilizes widely-used transformations including resizing, crop, rotation, color jitter, translation, and horizontal flip. 

\begin{figure*}[ht!]
\centering
\includegraphics[width=\linewidth]{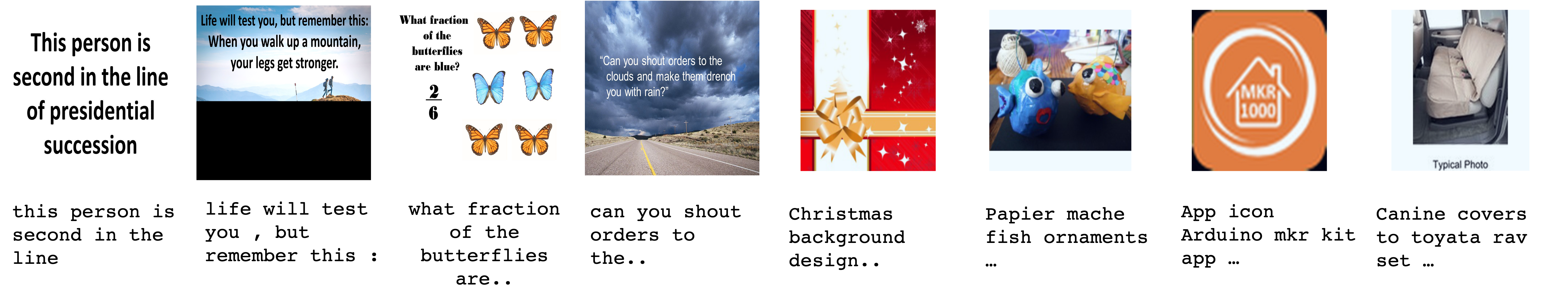}
\caption{Examples of non-member samples that receive high cosine similarity scores ($>0.3$). CSA cannot correctly mark these samples as non-members but WSA can.
}
\label{fig:memorize}
\end{figure*}

From Table~\ref{tab:main_laion}, we find that simple CSA already shows relatively high AUC scores (around $76\%$ on LAION and $68\%$ on CC12M). In addition, 
AEA consistently demonstrates slightly improved performance across all settings by $0.74\%$ to $7\%$.
While CSA and AEA demonstrate relatively high average performance, they fail in the low false-positive regime: the true positive rate is below $10\%$ at the false positive rate of $1\%$.
In contrast, WSA surpasses the performance of CSA and AEA in terms of all evaluation metrics by a significant margin. For instance, on the LAION dataset, WSA achieves an AUC of $93.2\%$ and TPR@1\%FPR of $74.01\%$, improving over CSA by $17\%$ in terms of AUC and $64\%$ in terms of TPR@1\%FPR. Figure~\ref{fig:memorize} illustrates examples that CSA fails to attack yet WSA succeeds. The effectiveness of WSA shows that there is a strong privacy implication of the one-sided non-member information available online. Overall, our results indicate a well-generalized multi-modal model trained with a large-scale dataset can still compromise privacy.

Interestingly, the attacks are less effective against models trained on CC12M than those trained on LAION. This is maybe because models trained on the CC12M dataset suffer more overfitting than models trained on the LAION dataset.
Moreover, given CC12M, the improvement induced by AEA is higher in RN50 and RN101 models compared with ViT-B/32. ViT-B/32 is overly large for the CC12M dataset, and the corresponding model suffers more overfitting than the other two ResNet-based architectures. 


\vspace{-0.5em}
\paragraph{In-depth Analysis of AEA.}
\begin{figure}[t!]
\centering
\includegraphics[width=1.0\linewidth]{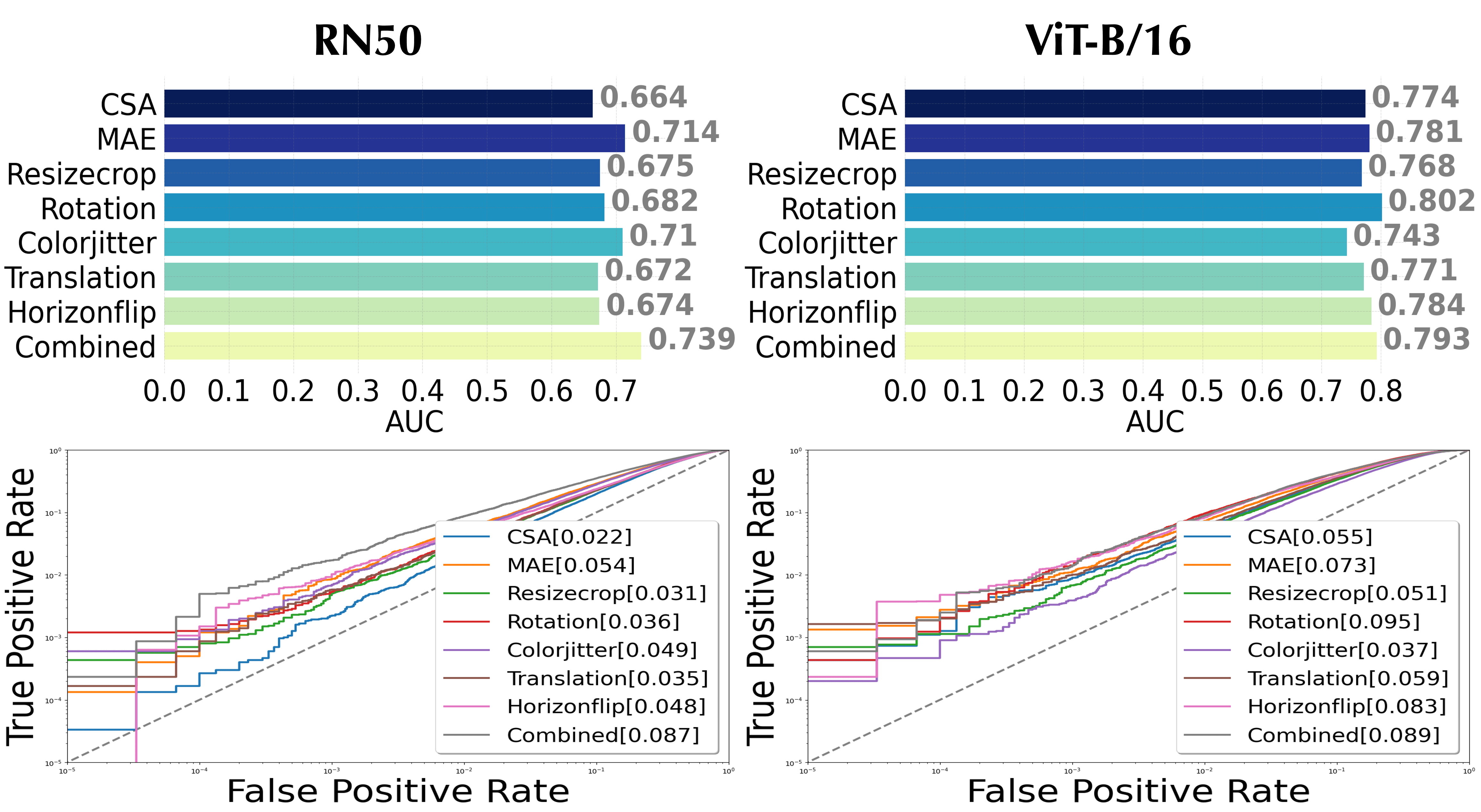}
\caption{AEA performance with various augmentations on models pre-trained on two different datasets. 
} 
\label{fig:aug_mix}
\end{figure}

We employ commonly utilized augmentations, such as resizing, crop, rotation, color jitter, translation, and horizontal flip, in conjunction with the masked autoencoder~\cite{he2022masked} to alter the feature space. In Figure~\ref{fig:aug_mix}, \texttt{Combined} denotes the AEA that utilizes all these augmentations. \texttt{Combined} is a reasonable strategy when the adversary does not have any knowledge to inform the selection of the augmentations. 

Our experiments, shown in Figure~\ref{fig:aug_mix}, evaluate the effectiveness of AEA across different choices of augmentations. 
Firstly, \texttt{Combined} outperforms the \texttt{CSA} baseline in terms of the AUC and TPR@1\%FPR metrics on both datasets, as demonstrated by the results provided in Table~\ref{tab:main_laion}. Moreover, \texttt{Combined} exhibits the best or the second best performance compared to using individual transformations. This indicates that without further information guiding the selection of transformations, using the \texttt{Combined} approach is recommended.

\begin{figure}[t!]
\centering
\includegraphics[width=\linewidth]{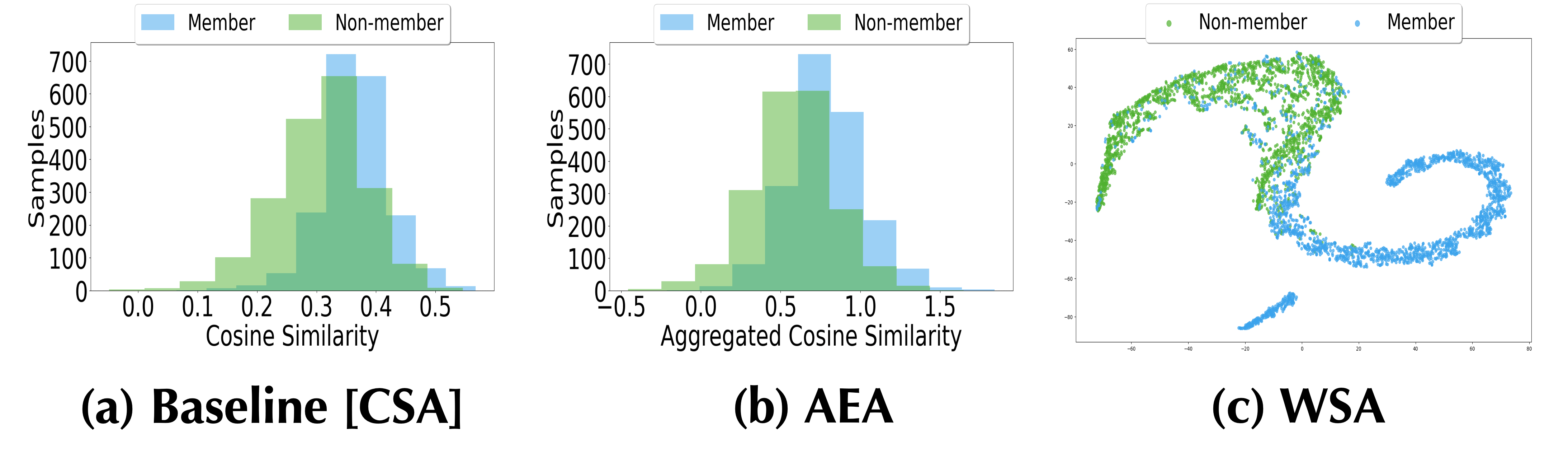}
\caption{Comparison of the separability between members and non-members among CSA, AEA, and WSA.
}
\label{fig:Mem_Nonmem}
\end{figure}

\vspace{-0.5em}
\paragraph{In-depth Analysis of WSA.} The cosine similarity score utilized by CSA can be regarded as a transformation of the image-text features but this transformation is too simple to extract the membership-indicative information and map members and non-members to distinctive clusters. On the other hand, WSA trains an attack model, which can be regarded as a more complex transformation, thereby more successfully extracting the membership-indicative information from the image-text features of the target point. As shown in Figure~\ref{fig:Mem_Nonmem}, it is difficult to achieve a clear separation between members and non-members based on the CS score (the metric of CSA) or the aggregate CS score (the metric of AEA). Yet, they are separable in \emph{some} latent space as shown in Figure~\ref{fig:Mem_Nonmem} (c) which projects the member and non-member features onto a two-dimensional embedding space found by t-SNE~\cite{van2008visualizing}. The efficacy of WSA is primarily attributed to its active discovery of a latent space where member and non-member samples can be effectively separated.


\begin{figure}[t!]
\centering
\includegraphics[width=1.0\linewidth]{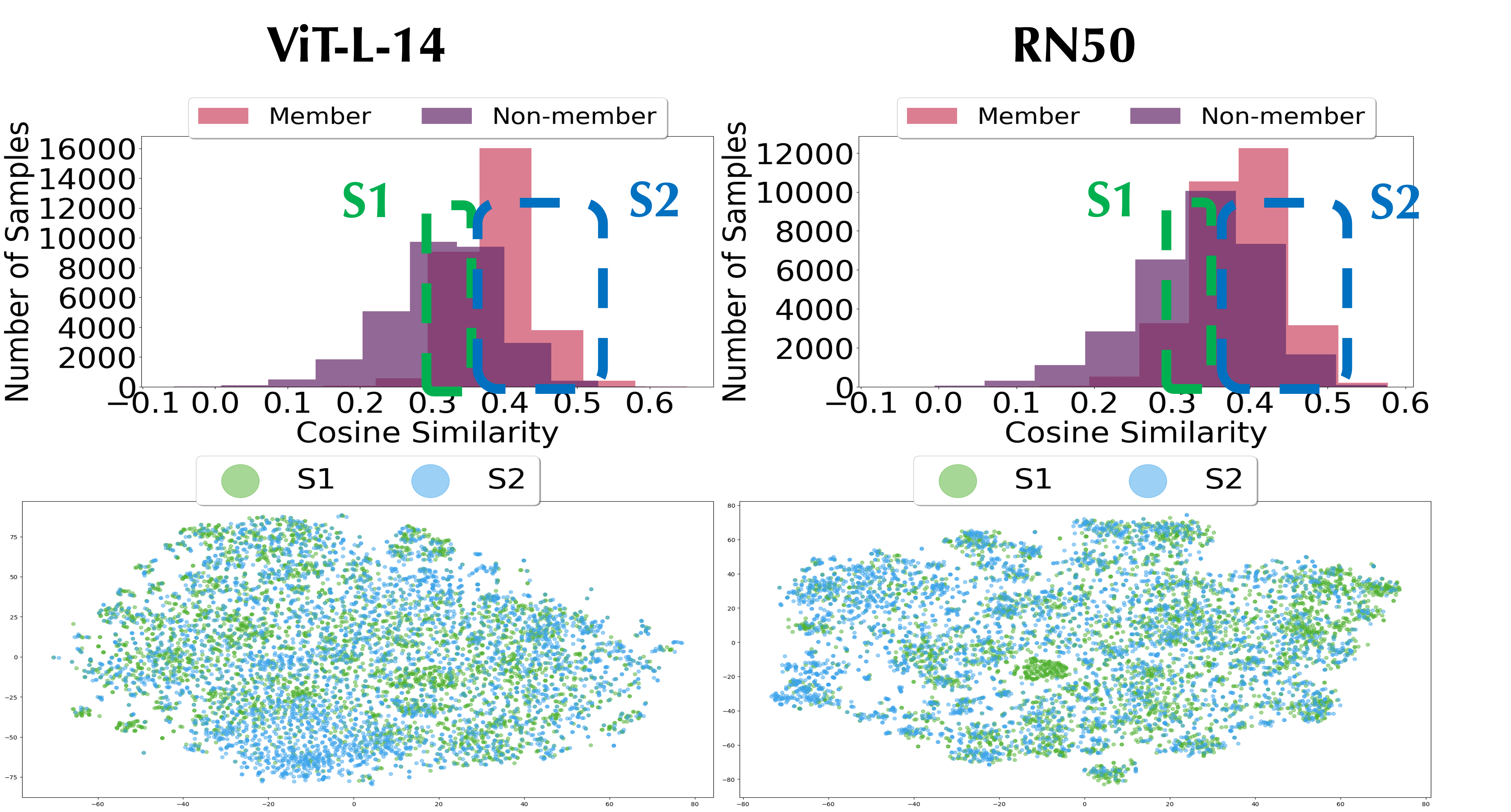}
\caption{T-SNE plots for features from the pseudo-labeled (S2) and the overlapped area (S1), illustrating the discriminative feature alignment between the two areas.}
\label{fig:tsne}
\end{figure}


Recall the pipeline of WSA (i.e., Figure ~\ref{fig:overview}). While the ground-truth non-member set has a diverse range of cosine similarity scores, the pseudo-members are labeled based on whether they have higher cosine similarity than the majority of the non-members. In other words, the pseudo-members are mostly from the area $S2$ in Figure~\ref{fig:tsne}. On the other hand, the members used for evaluation have a broad cosine similarity range: some members have large enough cosine similarity, yet others are indistinguishable from non-members using cosine similarity, and WSA achieves a high performance overall (over 92\% for all models trained with LAION and over 79\% for all models trained with CC12M). An intriguing question thus arises: \emph{Given that during training, the attack model has never seen members from the overlapped area (illustrated by $S1$ in Figure~\ref{fig:tsne}), how can the attack model learn to recognize these members?} 

We hypothesize that the effectiveness of WSA in the overlapped area is due to \textbf{discriminative feature alignment} between non-overlapped and overlapped areas, i.e., the members from both areas contain similar features indicative of their membership.
To verify the hypothesis, we randomly select data points whose CS scores lie in the overlapped area ($S1$) and conduct the same random sampling for the pseudo-labeled (or non-overlapped) area ($S2$). As shown in Figure~\ref{fig:tsne}, the features for members from the overlapped area have similar coverage to members from the non-overlapped area on both datasets. 
It is worth noting that features from the ViT-L/14 model, trained with the LAION dataset, exhibit better alignment compared to those from the RN50 model trained with CC12M. Specifically, the features extracted from the S2 area of the ViT-L/14 model span almost the entirety of the feature space; 
by contrast, the S2 features from RN50 are relatively sparse and concentrated and fail to cover the entire S1 area.
The findings regarding feature alignment shed light on the superior attack performance, in terms of TPR@1\%FPR, achieved by the LAION-trained ViT-L/14 model compared to the CC12M-trained RN50 model, as shown in Table~\ref{tab:main_laion}.
The samples from the overlapped area (S2) can be considered as \textit{hard-to-infer} samples, which contributes substantially to the TPR@1\%FPR, and discriminative feature alignment between S1 and S2 can increase the attack performance on \textit{hard-to-infer} samples.

\vspace{-0.5em}
\paragraph{Runtime.}
Since our main focus is on attacking large-scale models, we provide the runtime analysis of each attack method. Table~\ref{tab:main_laion} shows the results. CSA and AEA take 670.7s and 830.1s respectively. WSA takes 1539.2s to perform the attack. This time requirement is almost negligible compared to existing shadow training-based methods, which involve retraining CLIP-scale models many times. The result shows our proposed attach techniques are efficient and scalable to large-scale models in practice. 

\subsection{Sensitivity Analysis}
\label{sec:main_sensitivity}
We will delve into two factors that could potentially affect attack performance: 1) the size of accessible non-members $D_\text{no}$ and 2) the threshold for selecting pseudo-members.

\vspace{-0.5em}
\paragraph{Impact of $|D_\text{no}|$.}
We vary the size of $D_\text{no}$, ranging from 10K to 90K. 
and the results are summarized in Appendix ~\ref{sec:D}. The performance of WSA does vary with the size of accessible non-members yet only to a relatively small extent. In particular, it consistently outperforms the baseline for the range of $|D_\text{no}|$ considered. 

\begin{table*}[h!]
\caption{Attack performance vs. $\lambda$ and corresponding mislabeled pseudo-member ratio. 
$\lambda=0.5$ achieves the optimal performance.
}
\label{tab:mislabel}
\centering
\renewcommand{\arraystretch}{1.6}
\scalebox{0.60}{
\begin{tabular}{ccccccccccccc}
\hline
Dataset {[}Model{]} &
  \multicolumn{3}{c}{LAION {[}ViT-B-32{]}} &
  \multicolumn{3}{c}{LAION {[}ViT-L-14{]}} &
  \multicolumn{3}{c}{CC12M {[}ViT-B-32{]}} &
  \multicolumn{3}{c}{CC12M {[}RN50{]}} \\ \hline
Method &
  CSA &
  \multicolumn{2}{c}{WSA} &
  CSA &
  \multicolumn{2}{c}{WSA} &
  CSA &
  \multicolumn{2}{c}{WSA} &
  CSA &
  \multicolumn{2}{c}{WSA} \\ \hline
$\lambda$ {[}Mislabel ratio{]} &
  TPR@1\%FPR &
  TPR@1\%FPR &
  ACC &
  TPR@1\%FPR &
  TPR@1\%FPR &
  ACC &
  TPR@1\%FPR &
  TPR@1\%FPR &
  ACC &
  TPR@1\%FPR &
  TPR@1\%FPR &
  ACC \\ \hline
\textbf{Baseline} {[}37.53\%{]} &
  0.0671 &
  0.6973 &
  0.7390 &
  0.0456 &
  0.7072 &
  0.7376 &
  0.0400 &
  0.2911 &
  0.6475 &
  0.0401 &
  0.3132 &
  0.6631 \\
-1.5 {[}37.15\%{]} &
  0.0694 &
  0.6808 &
  0.7419 &
  0.0455 &
  \textbf{0.7178} &
  0.7781 &
  0.0440 &
  0.2966 &
  0.6559 &
  0.0233 &
  \textbf{0.3640} &
  0.6702 \\
-0.5 {[}34.92\%{]} &
  0.0688 &
  \textbf{0.7220} &
  0.7338 &
  0.0461 &
  0.7050 &
  0.7501 &
  0.0412 &
  \textbf{0.3132} &
  0.6698 &
  0.0225 &
  \textbf{0.3703} &
  \textbf{0.6949} \\
0 {[}32.87\%{]} &
  0.0636 &
  \textbf{0.7106} &
  \textbf{0.8027} &
  0.0475 &
  0.7149 &
  0.7675 &
  0.0392 &
  \textbf{0.3000} &
  0.6779 &
  0.0264 &
  0.3457 &
  0.6856 \\
0.5 {[}29.94\%{]} &
  0.0653 &
  0.6501 &
  \textbf{0.8081} &
  0.0450 &
  \textbf{0.7199} &
  \textbf{0.8135} &
  0.0419 &
  0.2733 &
  \textbf{0.6966} &
  0.0205 &
  0.3036 &
  \textbf{0.7029} \\
1.0 {[}26.43\%{]} &
  0.0671 &
  { 0.6304} &
  { 0.7899} &
  0.0448 &
  0.6668 &
  \textbf{0.8199} &
  0.0408 &
  0.2604 &
  \textbf{0.7050} &
  0.0269 &
  0.2614 &
  0.6625 \\
1.5 {[}21.60\%{]} &
  0.0623 &
  { 0.5377} &
  { 0.7810} &
  0.0446 &
  0.5628 &
  0.8063 &
  0.0378 &
  0.2133 &
  0.6757 &
  0.0228 &
  0.1947 &
  0.6532 \\ \hline
\end{tabular}
}
\end{table*}

\vspace{-0.5em}
\paragraph{Impact of Mislabeling Ratio of Pseudo-members.}
For WSA, we label data samples whose cosine similarities are higher than $\mu_\text{no}+\lambda \sigma_\text{no}$ as pseudo-members. Therefore, the selection of $\lambda$ is crucial. 
We vary $\lambda$ from $-1.5$ to $1.5$ and evaluate our attack performance correspondingly in Table~\ref{tab:mislabel}. We add another baseline that randomly selects samples from $\mathcal{D}_\text{all}$ and labels all the selected samples as members. This baseline does not rely on $\lambda$.
Opting for a high threshold value enables us to collect cleaner training data samples; however, it may provide less information regarding samples that fall within the uncertain area (i.e., S1). Conversely, lowering the threshold may result in obtaining more information on the uncertain area, which could increase the TPR@1\%FPR, but would simultaneously reduce the attack model's performance, as indicated in Table~\ref{tab:mislabel}. 
Therefore, it is essential to balance both metrics to achieve reasonable performance. 
Our empirical findings suggest that $\lambda=0.5$ yields reasonable performance. The performance drop at the very large threshold value of $1.5$ is due to the insufficiency of pseudo-members; at the same time, there is little discriminative feature alignment between the non-overlapped area and the overlapped area. An in-depth analysis of Table~\ref{tab:mislabel} is provided in Appendix ~\ref{sec:D}.
\section{Defenses}
\label{sec:main_defenses}

\begin{table*}[ht]
\caption{Attack performance mitigation according to L2 regularization and data augmentation on ViT-B/32 model trained with CC12M.}
\label{tab:defense_cc12m}
\centering
\renewcommand{\arraystretch}{1.3}
\scalebox{0.75}{
\begin{tabular}{cccclccclcccl}
\hline
ViT-B-32 & \multicolumn{4}{c}{Original} & \multicolumn{4}{c}{L2 {[}$\alpha = 0.001${]}} & \multicolumn{4}{c}{DA}               \\ \hline
Metric &
  AUC &
  TPR@1\%FPR &
  ACC &
  \multicolumn{1}{c}{Zeroshot} &
  AUC &
  TPR@1\%FPR &
  ACC &
  \multicolumn{1}{c}{Zeroshot} &
  AUC &
  TPR@1\%FPR &
  ACC &
  \multicolumn{1}{c}{Zeroshot} \\ \hline
CSA      & 0.6952  & 0.0456 & -      &  & 0.6874      & 0.0401     & -          &      & 0.6589 & 0.0330 & -      &           \\
AEA &
  0.7050 &
  0.0775 &
  \textbf{-} &
  \multicolumn{1}{c}{21.67} &
  0.6885 &
  0.0689 &
  - &
  \multicolumn{1}{c}{16.87} &
  0.6687 &
  0.0515 &
  - &
  \multicolumn{1}{c}{22.87} \\
WSA      & 0.7754  & 0.3016 & 0.6779 &  & 0.7463      & 0.2748     & 0.6567     &      & 0.7533 & 0.2583 & 0.6574 & \textbf{} \\ \hline
\end{tabular}
}
\end{table*}


\begin{table*}[h!]
\caption{Attack performance mitigation according to the injection of various magnitudes of Gaussian noise.}
\label{tab:defense_gaussian}
\centering
\renewcommand{\arraystretch}{1.3}
\scalebox{0.65}{
\begin{tabular}{cccccccccccccccccc}
\hline
              & Model  & \multicolumn{4}{c}{ViT-B/32}      & \multicolumn{4}{c}{ViT-B/16}      & \multicolumn{4}{c}{RN50}          & \multicolumn{4}{c}{RN101}         \\ \hline
$\sigma$      & Metric & AUC    & TPR    & ACC-L  & VAL-L  & AUC    & TPR    & ACC-L  & VAL-L  & AUC    & TPR    & ACC-L  & VAL-L  & AUC    & TPR    & ACC-L  & VAL-L  \\ \hline
              & CSA    & 0.7434 & 0.0632 &        &        & 0.7622 & 0.0506 &        &        & 0.6653 & 0.0226 &        &        & 0.6845 & 0.0300 &        &        \\
$\sigma=0.0$  & AEA    & 0.7663 & 0.1056 & 0.6022 & 1.5306 & 0.7961 & 0.0925 & 0.6705 & 1.4910 & 0.7425 & 0.0893 & 0.3589 & 2.7325 & 0.7277 & 0.0880 & 0.2272 & 2.1100 \\
              & WSA    & 0.9252 & 0.6768 &        &        & 0.9352 & 0.6655 &        &        & 0.8058 & 0.3087 &        &        & 0.8325 & 0.3927 &        &        \\ \hline
              & CSA    & 0.7354 & 0.0593 &        &        & 0.7567 & 0.0518 &        &        & 0.6624 & 0.0171 &        &        & 0.6739 & 0.0329 &        &        \\
$\sigma=0.01$ & AEA    & 0.7396 & 0.0901 & 0.6018 & 1.7223 & 0.7712 & 0.0853 & 0.6705 & 1.6410 & 0.7262 & 0.0780 & 0.3589 & 2.9098 & 0.215  & 0.0848 & 0.2274 & 2.1942 \\
              & WSA    & 0.9169 & 0.6510 &        &        & 0.9291 & 0.6677 &        &        & 0.8027 & 0.3138 &        &        & 0.8073 & 0.3421 &        &        \\ \hline
              & CSA    & 0.4979 & 0.0095 &        &        & 0.4991 & 0.0100 &        &        & 0.4962 & 0.0129 &        &        & 0.4976 & 0.0116 &        &        \\
$\sigma=0.5$  & AEA    & 0.5038 & 0.0098 & 0.4055 & 1706.2 & 0.4997 & 0.0110 & 0.5006 & 1697.2 & 0.4999 & 0.0103 & 0.2867 & 2252.4 & 0.5035 & 0.0121 & 0.2218 & 865.43 \\
              & WSA    & 0.4795 & 0.0097 &        &        & 0.4893 & 0.0085 &        &        & 0.4938 & 0.0081 &        &        & 0.4958 & 0.0087 &        &        \\ \hline
              & CSA    & 0.4971 & 0.0093 &        &        & 0.4982 & 0.0104 &        &        & 0.5051 & 0.0101 &        &        & 0.5033 & 0.0100 &        &        \\
$\sigma=1.0$  & AEA    & 0.5114 & 0.109  & 0.1204 & 6832.4 & 0.5054 & 0.0091 & 0.1711 & 6821.2 & 0.5050 & 0.0120 & 0.1534 & 9048.3 & 0.5134 & 0.0108 & 0.2094 & 3484.9 \\
              & WSA    & 0.4969 & 0.0087 &        &        & 0.4923 & 0.0098 &        &        & 0.4925 & 0.0109 &        &        & 0.4934 & 0.0094 &        &        \\ \hline
\end{tabular}
\vspace{-0.5em}
}
\end{table*}

While defense against MIAs is not the primary focus of this work, we explore several well-established defense strategies for simple models, such as regularization~\cite{truex2018towards}, data augmentation, and differential privacy~\cite{shokri2015privacy, abadi2016deep, jayaraman2019evaluating}, and re-examine their effectiveness in the context of large-scale multi-modal learning.

\vspace{-0.5em}
\paragraph{$L_2$ Regularization.}
We choose the regularization hyperparameter $\alpha = 0.001$,
as a higher value of $\alpha$ leads to non-convergence of loss.
Our results are summarized in Table~\ref{tab:defense_cc12m}. We observe that WSA is resilient to the regularization-based defense in all metrics ($0.7754 \rightarrow 0.7463$ for AUC, $0.3016 \rightarrow 0.2748$ for TPR@1\%FPR, and $0.6779 \rightarrow 0.6567$ for ACC). AEA performance also drops by a little. However, the zero-shot performance on ImageNet deteriorates concurrently. Hence, the integration of L2 regularization mitigates the risk of privacy to a small extent but may potentially impede the utility.

\vspace{-0.5em}
\paragraph{Data Augmentation.} In contrast, we find that data augmentation-based mitigation achieves a better tradeoff between privacy and utility. We leverage popular augmentations such as rotation, translation, and horizontal flip during training. The results in Table~\ref{tab:defense_cc12m} show that the AUC score for WSA drops from $0.7754$ to $0.7533$ (TPR@1\%FPR decreased by $0.0433$), with an increase of $1.2\%$ in zero-shot performance.  We additionally provide the results on the RN50 model trained on CC12M with L2 and data augmentation in Appendix ~\ref{sec:E}.

\vspace{-0.5em}
\paragraph{Feature Perturbation.}
Because of the high computational costs incurred by modifying the CLIP training process, we consider a simple defense of adding noise into the output features of a pre-trained CLIP model, which is in the same spirit as output perturbation approaches for classification models~\cite{jia2019memguard}. 
We inject zero-mean Gaussian noise with different standard deviations into the features. The results are shown in Table~\ref{tab:defense_gaussian}. We include two evaluation metrics, namely, the ImageNet classification accuracy (i.e., ACC-I) and the validation loss (i.e., VAL-L). Increasing the Gaussian noise scale from $\sigma=0$ to $\sigma=1.0$ reduces attack performance to random guessing (i.e., $0.5$). In particular, until $\sigma=0.01$, WSA, AEA, and CSA show relatively high attack performance. Hence, if we want to defend against WSA, noise with at least $\sigma=0.5$ is needed; however, at the same time, it causes a significant increase in validation loss and degradation in zero-shot performance (e.g., from $0.60$ to $0.40$ for ViT-B/32).   

\vspace{-0.5em}
\paragraph{Differential Privacy (DP).} 
\label{sec:dp} DP offers a formal guarantee for protecting individual records, which is suitable for defending against MIAs. We implement the DP with the Opacus library~\cite{opacus}. To reduce the dependency among gradients within a batch, we replace the batch normalization layer with an alternative normalization as suggested in Opacus. However, this substitution has already led to convergence issues of CLIP (see Appendix ~\ref{sec:E}).
Furthermore, training a large-scale model incorporating differential privacy leads to resource limitation issues.
Therefore, incorporating DP into CLIP necessitates significant additional efforts to re-design the contrastive training process to be DP-friendly (i.e., reducing the dependency between individual gradients) and computationally efficient. We will leave it as an open problem.

\section{Conclusion and Outlook} 

This paper presents the first focused study on application-agnostic MIAs against large-scale multi-modal models. To that end, we develop three strategies that make MIAs practical for large-scale models. Our evaluation shows the efficacy of the proposed attack strategies. Furthermore, our experimental findings emphasize the importance of safeguarding privacy for multi-modal foundational models, which were previously believed to be less susceptible to overfitting.


Our work opens several exciting directions for future work. First, it will be interesting to adapt our attack method to other foundation models, like BERT  and GPT-3. Second, while the injection of Gaussian noise provides a simple defense, it does not come with provable privacy guarantees. How to build high-utility foundation models with rigorous privacy guarantees is an interesting open question. 

\section{Acknowledgement}
RJ and the ReDS lab acknowledge support through a grant from the Amazon-Virginia Tech Initiative for Efficient and Robust Machine Learning.

\cleardoublepage
\newpage

{\small
\bibliographystyle{ieee_fullname}
\bibliography{ICCV_Camera_ready.bbl}

}

\cleardoublepage
\newpage
\appendix
\cleardoublepage
\newpage

\section{Additional Results on WSA [Section ~\ref{sec:main_eval}]}
\label{sec:B}
In this section, we additionally present the accuracy metric obtained from trained attack models in Table~\ref{tab:acc_attackmodel}. Given that it is infeasible to compute the accuracy values from CSA and AEA, we opted to showcase the AUC and TPR@1\%FPR in the main paper to ensure a fair comparison between the three approaches. Nevertheless, as a crucial design preference, it may be imperative to take into account the balanced scenarios between the true negative and true positive cases. Therefore, we furnish the outcomes in terms of accuracy and provide an explanation of the true negative and true positive analyses in Appendix~\ref{sec:D}. 


\begin{figure*}[h!]
\centering
\includegraphics[width=\linewidth]{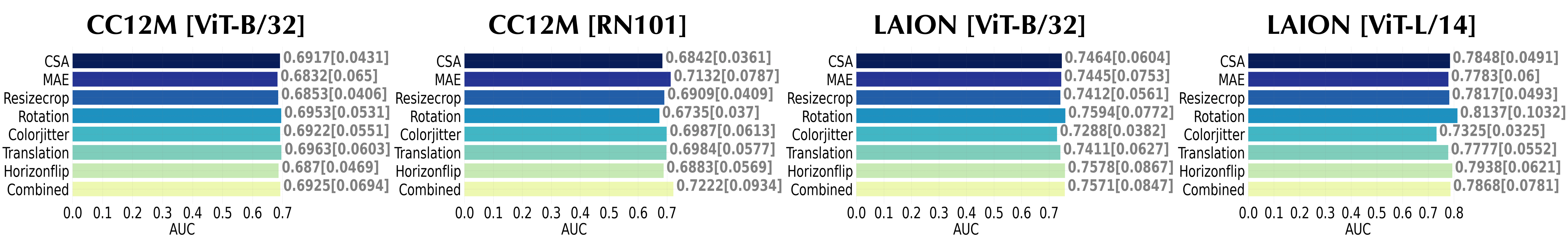}
\caption{AEA results on ViT-B/32 and ViT-L/14 trained with LAION and ViT-B/32 and RN101 trained with CC12M.
}
\label{fig:aug_additional}
\end{figure*}

\begin{table}[h!]
\caption{WSA performance in terms of accuracy obtained from attack models.} 
\label{tab:acc_attackmodel}
\centering
\renewcommand{\arraystretch}{1.7}
\scalebox{0.75}{
\begin{tabular}{ccc}
\hline
\multicolumn{2}{c}{{Method}}                      & {WSA}                     \\ \hline
{Dataset} & {Model} & {ACC}                     \\ \hline
                                        & {ViTi-B/32}      & {0.7845 $\pm$ 0.0122} \\
{LAION} & {ViTi-B/16} & {0.7996 $\pm$ 0.0062} \\
                                        & {ViTi-L/14}      & {0.8078 $\pm$ 0.0100} \\ \hline
                                        & {RN50}           & {0.6978 $\pm$ 0.0050} \\
{CC12M}            & {RN101}          & {0.7140 $\pm$ 0.0163} \\
                                        & {ViT-B/32}       & {0.7005 $\pm$ 0.0022} \\ \hline
\end{tabular}
}
\end{table}

\begin{figure}[h!]
\centering
\includegraphics[width=\linewidth]{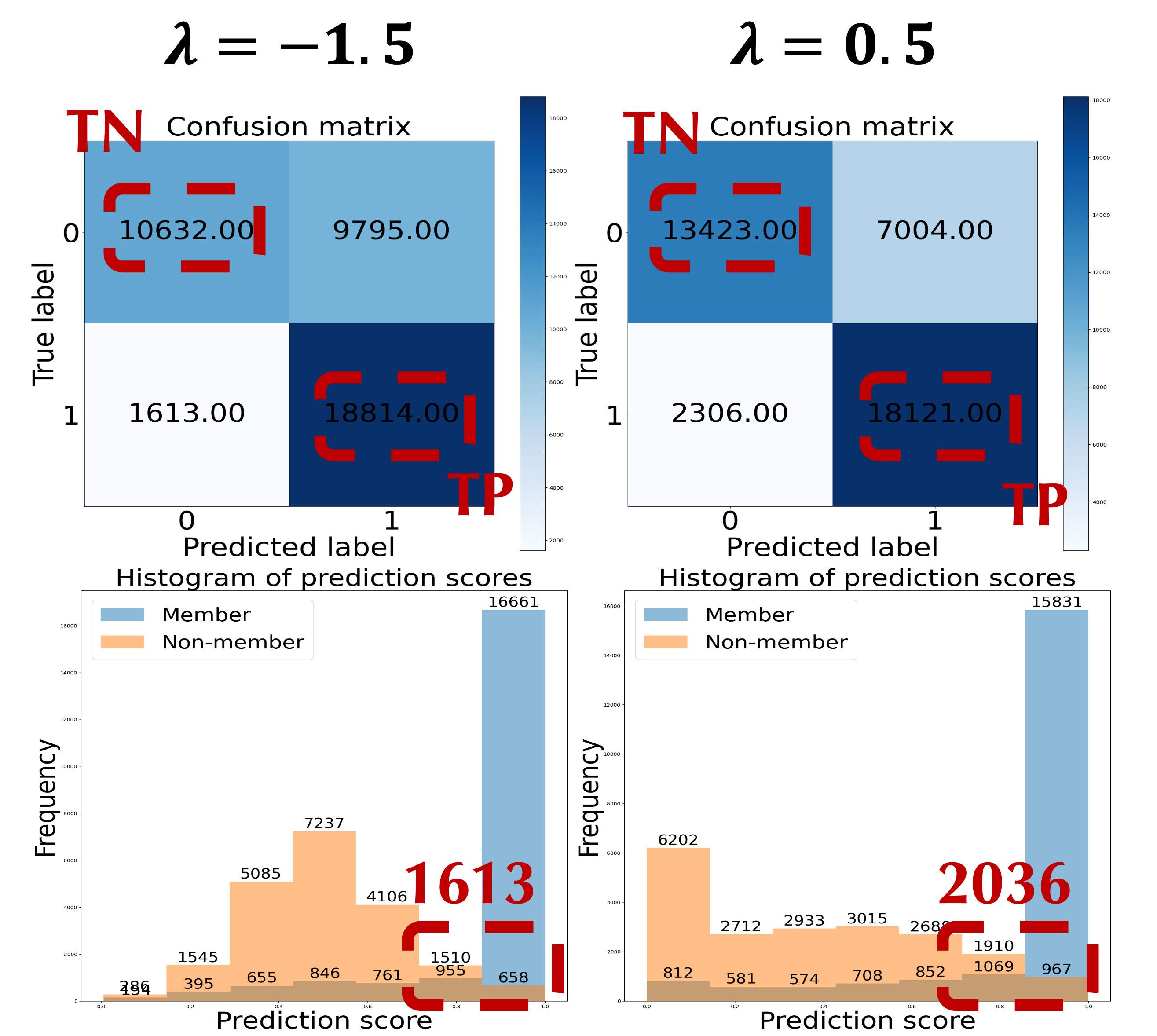}
\caption{Confusion matrix and histogram of prediction score analysis according to the different mislabel ratios. The confusion matrix serves to explicate the rationale underlying the observed decrease in accuracy scores, while concurrently elucidating the possibility of increased AUC scores through the augmentation of mislabeled data samples. Specifically, setting the threshold at $\lambda=-1.5$ leads to a marginal increase in the number of true positive (TP) cases, while concurrently causing a significant decrease in the number of true negative (TN) cases. These observations are consistent with the histogram plot, which demonstrates a gradual increase in the probability estimate of positive class on non-member samples ($\lambda=-1.5$) while decreasing the number of orange samples in the blue bar.} 
\label{fig:mislabel-analysis}
\end{figure}

\begin{table*}[t!]
\caption{WSA performance and the size of non-member data changes on LAION ViT-B/32 trained with LAION and RN50 trained with CC12M. When we set the non-member size as 70K, we can achieve the top-2 performance in three spots (i.e., TPR@1\%FPR, ACC on LAION, and TPR@1\%FPR on CC12M). However, even in case when the non-member samples are limited in size, WSA outperforms CSA.} 
\label{tab:nontrain}
\centering
\renewcommand{\arraystretch}{1.7}
\scalebox{0.8}{
\begin{tabular}{ccccccccccc}
\hline
Dataset {[}Model{]} & \multicolumn{5}{c}{LAION {[}ViT-B-32{]}}                                & \multicolumn{5}{c}{CC12M {[}RN50{]}}                   \\ \hline
Method              & \multicolumn{2}{c}{CSA}                  & \multicolumn{3}{c}{WSA}      & \multicolumn{2}{c}{CSA} & \multicolumn{3}{c}{WSA}      \\ \hline
Knowledge level     & AUC    & TPR@1\%FPR                      & AUC    & TPR@1\%FPR & ACC    & AUC      & TPR@1\%FPR   & AUC    & TPR@1\%FPR & ACC    \\ \hline
10K                 & 0.7439 & 0.06556 & 0.8841 & 0.5852     & 0.7691 & 0.6854   & 0.0366       & 0.7931 & 0.3123     & 0.6757 \\
30K                 & 0.7439 & 0.06556 & 0.9024 & 0.6501     & 0.8081 & 0.6854   & 0.0366       & 0.7860 & 0.2733     & 0.6966 \\
50K                 & 0.7439 & 0.06556 & 0.9102 & 0.6811     & 0.7842 & 0.6854   & 0.0366       & 0.7925 & 0.2944     & 0.6948 \\
70K                 & 0.7439 & 0.06556 & 0.9216 & 0.7337     & 0.8011 & 0.6854   & 0.0366       & 0.7855 & 0.3143     & 0.6910 \\
90K                 & 0.7439 & 0.06556 & 0.9074 & 0.6974     & 0.7725 & 0.6854   & 0.0366       & 0.7894 & 0.3444     & 0.6928 \\ \hline
\end{tabular}
}
\end{table*}

\begin{table*}[h!]
\caption{Attack performance mitigation according to L2 regularization and data augmentation on RN50 model trained with CC12M.}
\label{tab:app_defenses}
\centering
\renewcommand{\arraystretch}{1.7}
\scalebox{0.75}{
\begin{tabular}{cccclccclcccl}
\hline
RN50 &
  \multicolumn{4}{c}{Self-trained Model} &
  \multicolumn{4}{c}{{L2 {[}$\alpha = 0.001${]} }} &
  \multicolumn{4}{c}{{DA}} \\ \hline
Metric &
  AUC &
  TPR@1\%FPR &
  ACC &
  \multicolumn{1}{c}{Zeroshot} &
  { AUC} &
  { TPR@1\%FPR} &
  { ACC} &
  \multicolumn{1}{c}{{ Zeroshot}} &
  { AUC} &
  { TPR@1\%FPR} &
  { ACC} &
  \multicolumn{1}{c}{Zeroshot} \\ \hline
CSA &
  0.7861 &
  0.0322 &
  - &
   &
  { 0.6770} &
  { 0.0363} &
  { -} &
  { } &
  { 0.7835} &
  { 0.0391} &
  { -} &
   \\
AEA &
  0.8103 &
  0.1218 &
  - &
  0.3100 &
  { 0.7037} &
  { 0.0965} &
  { -} &
  { 0.1366} &
  { 0.8058} &
  { 0.1058} &
  { -} &
  0.3217 \\
WSA &
  0.8839 &
  0.4587 &
  0.7813 &
   &
  { 0.7919} &
  { 0.3446} &
  { 0.6989} &
  { } &
  { 0.8758} &
  { 0.4320} &
  { 0.7719} &
   \\ \hline
\end{tabular}
}
\end{table*}

\begin{figure}[h!]
\centering
\includegraphics[width=0.9\linewidth]{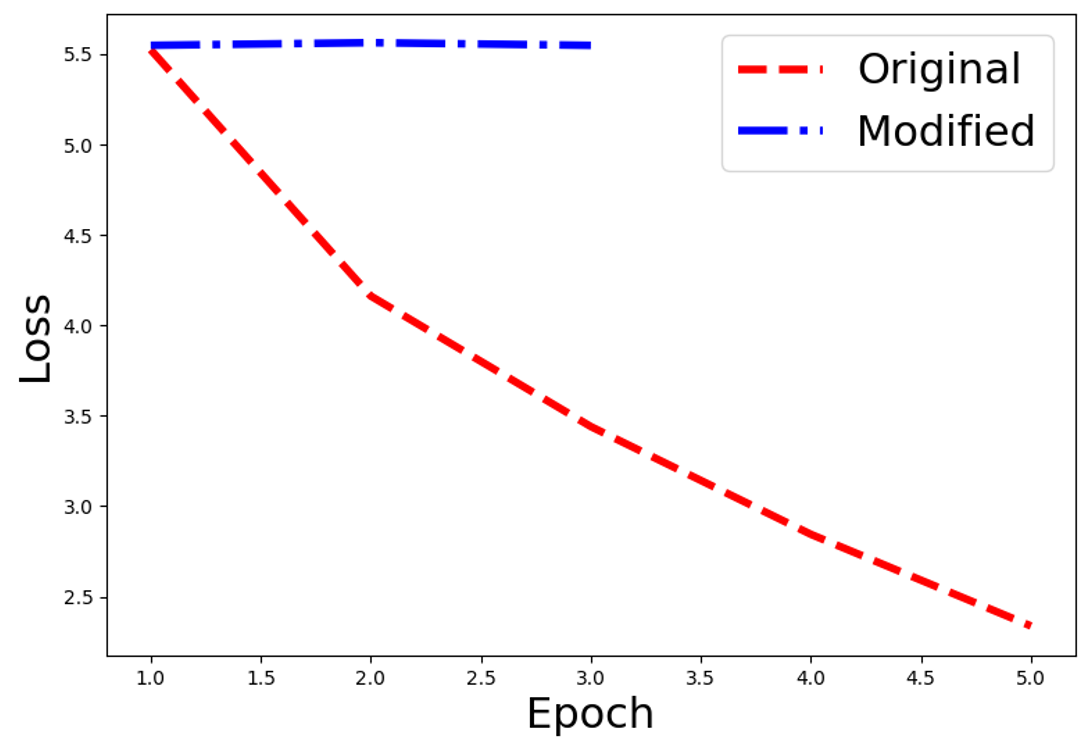}
\caption{Loss convergence according to the original model and modified model (i.e., a model obtained from ~\cite{opacus}).}
\vspace{-0.5em}
\label{fig:DPloss}
\end{figure}

\section{Additional Results and Analysis on AEA [Section ~\ref{sec:main_eval}]}
\label{sec:C}

We additionally present  AEA results on different models trained with different datasets (e.g., RN101, ViT-B/32 with CC12M and ViT-B/32, ViT-L/14 with LAION) in Figure~\ref{fig:aug_additional}. As shown in the figure, AEA surpasses CSA in all models trained with different datasets in terms of AUC and TPR@1\%FPR. 

Regarding the performance of AEA, we observe that on the pre-trained models trained with LAION, AEA exhibits the best performance with rotation and the worst performance with colorjitter. By contrast, for the RN50, and RN101 trained with CC12M, AEA performs best or second best with colorjitter and Masked Autoencoder (MAE) augmentations, except for the combined augmentation. These results suggest that pre-trained models with the LAION dataset (e.g., LAION ViT-B/16, ViT-B/32, and ViT-L/14) show robustness towards colorjitter, but are weak to rotation changes, while self-trained models with CC12M exhibit a weakness towards MAE and colorjitter.

\section{Sensitivity Analysis on $|D_\text{no}|$ and mislabel ratio [Section ~\ref{sec:main_sensitivity}]}
\label{sec:D}

\paragraph{In-depth Analysis on the Impact of $|D_\text{no}|$.}
Attack performance in terms of ACC increases until $|D_\text{no}| = 70K$ and decreases. 
The reason is that mere expansion of the size leads to including more noisy samples without providing useful alignment information between samples, which leads to a decrease in performance.
Furthermore, regardless of the size of non-training data, our proposed method, WSA, consistently outperforms the baseline for both datasets and two different models.

\paragraph{In-depth Analysis on the Impact of Mislabeling Ratio.}
A comprehensive analysis of the results from Table~\ref{tab:mislabel} may raise the question of why TPR@1\%FPR sometimes shows better performance with lower accuracy. For example, for the LAION ViT-L/14 pre-trained model, TPR@1\%FPR is $0.7178$ when $\lambda=-1.5$, but TPR@1\%FPR drops to $0.6668$ with higher attack model accuracy (i.e., $0.8199$). Therefore, we provide analysis regarding the performance increase in terms of the TPR@1\%FPR at the low threshold value even with the high mislabeled ratio. 
As depicted in Figure~\ref{fig:mislabel-analysis}, selecting a lower threshold value (i.e., $\lambda=-1.5$) results in a less number of non-member samples at high confidence values (e.g., $2036 \rightarrow 1613$). This, in turn, leads to an increase in the number of true positive (TP) cases as described in the confusion matrices (e.g., $18121 \rightarrow 18814$). However, this advantage comes at the cost of sacrificing true negative (TN) simultaneously (e.g., $13423 \rightarrow 10632$). It becomes more challenging to correctly classify non-members, thus lowering the ACC score.
In addition, as presented in the table, the baseline approach fails to achieve high accuracy even though it achieves relatively high TPR@1\%FPR. In case attackers prioritize TPR@1\%FPR, they may opt for a lower threshold. 
In sum, WSA provides superior performance, compared to CSA, for any threshold selection. Nevertheless, by carefully selecting $\lambda$, we can achieve a more balanced performance.

\section{Additional Results on Defenses [Section ~\ref{sec:main_defenses}]}
\label{sec:E}
In this section, we further present the defense results obtained from the RN50 models with data augmentation and L2 regularization. Even though ~\cite{cherti2022reproducible} provides the pre-trained RN50 model on CC12M, to fairly compare the results, we train three RN50 models from scratch on CC12M (i.e., a model without defense, a model with DA, a model with L2 regularization). We use similar hyperparameter settings provided in the original paper. In particular, we set the number of epochs to 30, a learning rate to $1e-3$, and a weight decay for DA and original models to $1e-1$. The zero-shot accuracy on ImageNet is near $31\%$ for the self-trained original RN50 model, and the accuracy from the pre-trained model is around $35\%$. The defense results are summarized in Table~\ref{tab:app_defenses}.

\paragraph{$L_2$ Regularization} As described in Table~\ref{tab:app_defenses}, similar to results from Section ~\ref{sec:main_defenses}, we find that L2 regularization is effective to curtail the attack performance in terms of all metrics. In particular, the AUC score for WSA drops from 0.8839 to 0.7919 and the TPR@1\%FPR also exhibits a decrease from 0.4587 to 0.3446. Similarly, the AEA is mitigated (e.g., the decrease in AUC score by 0.1066 and TPR@1\%FPR by 0.0253).
However, this attack mitigation comes at a cost of utility degradation (i.e., zeroshot performance).

\paragraph{Data Augmentation} Our findings suggest that Data Augmentation (DA) is still capable of providing effective defense results while simultaneously improving utility for the RN50 model. However, the degree of attack mitigation achieved is insignificant. Specifically, DA results in a slight increase of 0.0117 in the zero-shot performance, while simultaneously resulting in a decrease in AUC score by 0.0081 and TPR@1\%FPR by 0.0267 for WSA. Moreover, DA decreases the AUC score by 0.0045 and TPR@1\%FPR by 0.016. 

\paragraph{Differential Privacy}
We continue our discussion in Section~\ref{sec:dp}. Since the batch-normalization layer cannot provide the privacy guarantee, we simply replace the corresponding layers with the layers suggested by ~\cite{opacus} and show the loss on each epoch during training in Figure~\ref{fig:DPloss}. 

As depicted in the figure, it is apparent that the modified model, which is obtained by replacing the layers  without incorporating the DP algorithm (e.g., adding noise, and gradient clipping), fails to attain loss convergence. Conversely, the original model from ~\cite{cherti2022reproducible} yields the loss convergence. In this experiment, we leverage 600K image and text pairs on the RN50 vision encoder. We note that after the third epoch, the loss for the DP model goes to Nan. Therefore, it is necessary to investigate how to properly incorporate the DP algorithm into large-scale multi-modal training in future work.

\section{Data Processing [Section ~\ref{sec:main_data}]}
\label{sec:F}

As described in the main paper, since most of the data samples are scraped from the Internet, there is an overlap between datasets. To consider $\mathcal{D}_\text{no}$ and separate $D_\text{attack}$ from $D_\text{eval}$ for the attack model, it is important to check the overlapping pairs between datasets. 

In this experiment, we adopt commonly used text pre-processing steps: 1) remove spacing, 2) lowering, 3) remove numbers, 4) remove punctuation, and 5) remove stopwords. After processing all captions, we exclude the common pairs between considered sets to meet $\mathcal{D}_\text{no} \cup D_\text{trn} = \emptyset$ and $D_\text{attack} \cup D_\text{eval} = \emptyset$. We additionally check the URL overlap for the images.  



\end{document}